\documentclass{article}
\usepackage{arxiv}

\usepackage[utf8x]{inputenc}
\usepackage{hyperref}       % hyperlinks
\usepackage{url}            % simple URL typesetting
\usepackage{booktabs}
\usepackage{amsfonts}
\usepackage{microtype}
\usepackage{import}
\usepackage{amsmath}
\usepackage{float}
\usepackage{multirow}
\usepackage{colortbl}
\usepackage{longtable}
\usepackage{authblk}
\usepackage{adjustbox}
\usepackage[framemethod=default]{mdframed}

\usepackage{caption} 
\usepackage[thai,english]{babel}
\usepackage{fonts-tlwg}

\mdfdefinestyle{default}{%
rightline=true,
innerleftmargin=10,
innerrightmargin=10,
frametitlerule=true,
frametitlerulecolor=green,
frametitlebackgroundcolor=black,
frametitlerulewidth=1pt}

\title{WangchanBERTa: Pretraining transformer-based Thai Language Models}

\author[  \dag]{\textbf{Lalita Lowphansirikul}\thanks{Equal contribution. Listed in random order.} }

\author[ \ddag]{\textbf{Charin Polpanumas}\textsuperscript{$*$}}

\author[ \dag]{\textbf{Nawat Jantrakulchai}}
\author[ \dag]{\textbf{Sarana Nutanong}}

\affil[ \ddag]{PyThaiNLP} \affil[ ]{\textup{charin.polpanumas@datatouille.org}}

\affil[ \dag]{School of Information Science and Technology, Vidyasirimedhi Institution of Science and Technology}
\affil[ ]{\textup{\{lalital\_pro, nawatj\_pro, snutanon\}@vistec.ac.th}}

\begin{document}
\maketitle
\begin{abstract}

Transformer-based language models, more specifically BERT-based architectures have achieved state-of-the-art performance in many downstream tasks. However, for a relatively low-resource language such as Thai, the choices of models are limited to training a BERT-based model based on a much smaller dataset or finetuning multi-lingual models, both of which yield suboptimal downstream performance. Moreover, large-scale multi-lingual pretraining does not take into account language-specific features for Thai. To overcome these limitations, we pretrain a language model based on RoBERTa-base architecture on a large, deduplicated, cleaned training set (78GB in total size), curated from diverse domains of social media posts, news articles and other publicly available datasets. We apply text processing rules that are specific to Thai most importantly preserving spaces, which are important chunk and sentence boundaries in Thai before subword tokenization. We also experiment with word-level, syllable-level and SentencePiece tokenization with a smaller dataset to explore the effects on tokenization on downstream performance. Our model wangchanberta-base-att-spm-uncased trained on the 78.5GB dataset outperforms strong baselines (NBSVM, CRF and ULMFit) and multi-lingual models (XLMR and mBERT) on both sequence classification and token classification tasks in human-annotated, mono-lingual contexts. 

\end{abstract}

\section{Introduction}

% First Paragraph
% - Give a very broad context of what LMs are used for
% - Conclude the paragraph with what we want we need to do effectively support the aforementioned downstream task: (i) large, deduplicated, and cleaned training data; (ii) diverse domains - social media, news, open datasets; (iii) space-aware subword tokenizer

Transformer-based language models, more specifically BERT-based architectures \cite{devlin2018bert}, \cite{liu1907roberta}, \cite{lan2019albert}, \cite{clark2020electra}, and \cite{he2020deberta}, have achieved state-of-the-art performance in downstream tasks such as sequence classification , token classification, question answering, natural language inference and word sense disambiguation \cite{wang2018glue,wang2019superglue}. However, for a relatively low-resource language such as Thai, the choices of models are limited to training a BERT-based model based on a much smaller dataset such as BERT-th \cite{thaikerasbert} trained on \textit{Thai Wikipedia Dump}, or finetuning multi-lingual models such as XLMR \cite{conneau2019unsupervised} (100 languages) and mBERT \cite{devlin2018bert} (104 languages). Training on a small dataset has a detrimental effect on downstream performance. BERT-th underperforms RNN-based ULMFit \cite{charin_polpanumas_2021_4429691} trained \textit{Thai Wikipedia Dump} on sequence classification task \textit{Wongnai Reviews} \cite{wongnai2018}. For multi-lingual training, we can see from comparison between multi-lingual and mono-lingual models such as \cite{martin2019camembert} that multi-lingual models underperform mono-lingual models. Moreover, large-scale multi-lingual pretraining does not take into account language-specific features for Thai. To overcome these limitations, we pretrain a language model based on RoBERTa-base architecture on a large, deduplicated, cleaned training set (78GB in total  size), curated from diverse domains of social media posts, news articles and other publicly available datasets. We apply text processing rules that are specific to Thai most importantly preserving spaces, which are important chunk and sentence boundaries in Thai before subword tokenization.

% Second Paragraph
%
% Pretty much the what you have currently. But I would start it with "In this report, we describe ...
%
In this report, we describe a language model based on RoBERTa-base architecture and SentencePiece \cite{sentencepiece} subword tokenizer on 78GB cleaned and deduplicated data from publicly available social media posts, news articles, and other open datasets. We also pretrain four other language models using different tokenizers, namely SentencePiece \cite{sentencepiece}, dictionary-based word-level and syllable-level tokenizer (PyThaiNLP's newmm \cite{pythainlp}), and SEFR tokenizer \cite{limkonchotiwat2020domain}, on Thai Wikipedia Dump to explore how tokens affect downstream performance.

% Third Paragraph
%
% 
To assess the effectiveness of our language model, we conducted an extensive set of experimental studies on the following downstream tasks: sequence classification (multi-class and multi-label) and token classification. Our model wangchanberta-base-att-spm-uncased outperforms strong baseline models (NBSVM \cite{wang2012baselines} and CRF \cite{okazaki2007crfsuite}), ULMFit \cite{howard2018universal} (thai2fit \cite{charin_polpanumas_2021_4429691}) and multi-lingual transformer-based models (XLMR \cite{conneau2019unsupervised} and mBERT \cite{DBLP:journals/corr/abs-1810-04805}) on both sequence and token classification tasks.

% Fourth Paragraph
%
The remaining sections of this report are organized as follows. In Section 2, we describe the methodology in pretraining the language models including raw data, preprocessing, train-validation-test split preparation and training the models. In Section 3, we introduce the downstream tasks we use to test the performance of our language models. In Section 4, we demonstrate the results of our language modeling and finetuning for downstream tasks. In Section 5, we discuss the results and next steps for this work.

%% added
The pretrained language models and finetuned models\footnote{https://huggingface.co/airesearch} are publicly available at  Huggingface's Model Hub. The source code used for the experiments can be found at our GitHub repository.\footnote{https://github.com/vistec-AI/thai2transformers}

\section{Methodology}

We train one language model on the \textit{Assorted Thai Texts dataset} including all available raw datasets and four language models on the \textit{Wikipedia-only dataset}, each with a different tokenizer.

\subsection{Raw Data}

The raw data are obtained from (statistics after preprocessing):

\renewcommand*{\arraystretch}{1.25}
\begin{longtable}{| >{\centering\arraybackslash} m{2.6cm} | >{\centering\arraybackslash} m{1.9cm} | >{\arraybackslash} m{10.7cm} |}

     \hline
     Dataset name & Data size & Description  \\  
     \hline 
     
     wisesight-large & \shortstack{51.44GB \\ (314M lines)} & a large dataset of social media posts provided by the social listening platform  Wisesight\footnote{https://wisesight.com/} for this study. The dataset contains posts Twitter, Facebook, Pantip, Instagram, YouTube and other websites sampled from 2019. \\ \hline
     
     pantip-large   &  \shortstack{22.35GB \\ (95M lines)} & a collection of posts and answers of Thailand's largest online bulletin board Pantip.com from 2015 to 2019 provided by audience analytics platform Chaos Theory.\footnote{https://www.facebook.com/ChaosTheoryCompany/}  \\ \hline
    
     Thairath-222k\footnote{https://github.com/nakhunchumpolsathien/TR-TPBS} &  
     \shortstack{1.48GB \\ (5M lines)} &   a collection of articles published on newspaper website Thairath.com up to December 2019. \\ \hline
     prachathai-67k\footnote{https://github.com/PyThaiNLP/prachathai-67k}   & 
     \shortstack{903.1MB \\ (2.7M lines)} &  a collection of articles published on newspaper website Prachathai.com from August 24, 2004 to November 15, 2018. \\ \hline

     Thai Wikipedia Dump\footnote{https://dumps.wikimedia.org/backup-index.html}  & \shortstack{515MB \\ (843k lines)}  &  the Wikipedia articles extracted using Giuseppe Attardi’s WikiExtractor\footnote{https://github.com/attardi/wikiextractor/} in September 2020. All HTML tags, bullet points, and tables are removed. 
     \\ \hline
     
     OpenSubtitles &  \shortstack{468.8MB \\ (5M lines)} & a collection of movie subtitles translated by crowdsourcing from OpenSubtitles.org \cite{opensubtitles2016}. We use only the portions containing Thai texts. \\ \hline
     
     ThaiPBS-111k\footnote{https://github.com/nakhunchumpolsathien/TR-TPBS}  & \shortstack{372.3MB \\ (858k lines) } & a collection of articles published on newspaper website ThaiPBS.or.th up to December 2019. \\ \hline
     
     Thai National Corpus & \shortstack{366MB \\ (797k lines)} & a 14-million-word corpus of Thai texts containing 75\% non-fiction and 25\% fiction works. Media source breakdown is 60\% books, 25\% magazines, and the rest from other publications and writings. Most of the texts are curated from 1998 to 2007 \cite{aroonmanakun2009thai}. \\ \hline
     
     scb-mt-en-th-2020 & \shortstack{290.4MB \\ (947k lines)}  & a parallel corpus of Englsih-Thai sentence pairs curated news, Wikipedia articles, SMS messages, task-based dialogs, web-crawled data, government documents, and machine-generated text \cite{lowphansirikul2020scb}. \\ \hline
     
     JW300 & \shortstack{ \\ 182.8MB \\ (727k lines)}  & a parallel corpus of religion texts from jw.org that includes Thai texts. \\ \hline
     
     wongnai-corpus\footnote{https://github.com/wongnai/wongnai-corpus} & \shortstack{64MB \\ (101k lines)} &  a collection of restaurant reivews and ratings (1 to 5 stars) published on Wongnai.com. \\ \hline
     
     QED & \shortstack{42MB \\ (407k lines) } & a collection of transcripts for educational videos and lectures collaboratively created on the AMARA web-based platform \cite{QED}. \\ \hline
     
     bibleuedin &  \shortstack{2.18MB \\ (62k lines)} & a multilingual corpus of the Bible created by Christos Christodoulopoulos and Mark Steedman. \\ \hline
     
     wisesight-sentiment & \shortstack{5.3MB \\ (22k lines)}  &  a collection of Twitter posts about consumer products and services from 2016 to early 2019 labeled positive, negative, neutral and question  \cite{bact_2019_3457447}. \\ \hline
     
     tanzil & \shortstack{2.4MB \\ (6k lines)} & a collection of Quran translations compiled by the Tanzil project \cite{TIEDEMANN12.463}. \\ \hline
     
     tatoeba & \shortstack{1MB \\ (2k lines)} & a collection of translated sentences from the crowdsourced multilingual dataset Tatoeba \cite{TIEDEMANN12.463}. \\ 
     \hline
\end{longtable}

\subsection{Preprocessing} \label{subsection:text_preprocessing}

\par We apply preprocessing rules to the raw datasets before using them as our training sets. This effectively demands the preprocessing rules to be applied before finetuning for both domain-specific language modeling and other downstream tasks.

\paragraph{Text Processing} A large portion of our training data (\textit{wisesight-large} and \textit{pantip-large}) comes from social media, which usually have a lot of unusual spellings and repetitions. For such noisy data, \cite{raffel2020exploring} reports that pretraining on a cleaned corpus \textit{C4} yields better performance in downstream tasks. Therefore, we opted to perform the following processing rules, in order:

\begin{itemize}
    \item Replace HTML forms of characters with the actual characters such as \textit{nbsp;} with a space and \textit{\foreignlanguage{thai}{<br />}} with a line break \cite{howard2018universal}.
    \item Remove empty brackets (\textit{()}, \textit{\{\}}, and \textit{[]}) than sometimes come up as a result of text extraction such as from Wikipedia.
    \item Replace line breaks with spaces.
    \item Replace more than one spaces with a single space
    \item Remove more than 3 repetitive characters such as \foreignlanguage{thai}{ดีมากกก} to \foreignlanguage{thai}{ดีมาก} \cite{howard2018universal}.
    \item Word-level tokenization using \cite{pythainlp}'s \textit{newmm} dictionary-based maximal matching tokenizer.
    \item Replace repetitive words; this is done post-tokenization unlike \cite{howard2018universal} since there is no delimitation by space in Thai as in English.
    \item Replace spaces with \textit{\foreignlanguage{thai}{<\_>}}. The SentencePiece tokenizer combines the spaces with other tokens. Since spaces serve as punctuation in Thai such as sentence boundaries similar to periods in English, combining it with other tokens will omit an important feature for tasks such as word tokenization and sentence breaking. Therefore, we opt to explicitly mark spaces with \textit{\foreignlanguage{thai}{<\_>}}. 
\end{itemize}

\par For \textit{Wikipedia-only dataset}, we only replace non-breaking spaces with spaces, remove an empty parenthesis that occur right after the title of the first paragraph, and replace spaces with \textit{\foreignlanguage{thai}{<\_>}}.

\paragraph{Sentence Breaking}
Each row of all datasets are originally delimited by line breaks. Due to memory constraints, in order to train the language model, we need to limit our maximum sequence length to 416 subword tokens (tokenized by SentencePiece \cite{sentencepiece} unigram model) or roughly 300 word tokens (tokenized by dictionary-based maximal matching \cite{pythainlp}). In order to do so, we use the sentence breaking model CRFCut (\cite{lowphansirikul2020scb}). CRFCut is a conditional random fields (CRF) model trained on English-to-Thai translated texts of \cite{sornlertlamvanich1997orchid} (23,125 sentences), TED transcripts (136,463 sentences; \cite{lowphansirikul2020scb}) and generated product reviews (217,482 sentences; \cite{lowphansirikul2020scb}). It uses English sentence boundary as sentence boundary labels for translated Thai texts. CRFCut has sentence-boundary F1 score of 0.69 on \cite{sornlertlamvanich1997orchid}, 0.71 on TED Transcripts, and 0.96 on generated product reviews. We keep only sentences that are 5 to 300 words long to not exceed 416-subword maximum sequence length and also not have a sequence too short for language modeling.

\paragraph{Tokenizers} For the model trained on \textit{Assorted Thai Texts dataset}, in the same manner as \cite{martin2019camembert}, we use SentencePiece \cite{sentencepiece} unigram language model \cite{kudo2018subword} to tokenize sentences of training data into subwords. The tokenizer has a vocabulary size of 25,000 subwords, trained on 15M sentences. To construct the training set for the tokenizer, we first take 2.5M randomly sampled sentences from \textit{pantip-large}, 3.5M randomly sampled sentences from \textit{wisesight-large} and all sentences of the remaining datasets, resulting in 20,961,306 total sentences. Out of those, we randomly sampled 15M  sentences to train the tokenizer.

For the models trained on \textit{Wikipedia-only dataset}, we use four different tokenizers to examine their effects on language modeling and downstream tasks. We use the same training set of 944,782 sentences sampled from \textit{Thai Wikipedia Dump}

\begin{itemize}
    \item \textbf{SentencePiece tokenizer}; we train the SentencePiece \cite{sentencepiece} unigram language model \cite{kudo2018subword} using 944,782 sentences from \textit{Thai Wikipedia Dump}, resulting in a tokenizer with vocab size of 24,000 subwords.
    \item \textbf{Word-level tokenizer}; the word-level, dictionary-based tokenizer \textit{newmm} \cite{pythainlp} is used to create a tokenizer with vocab size of 97,982 words.
    \item \textbf{Syllable-level tokenizer}; the syllable-level dictionary-based tokenizer \textit{syllable} \cite{pythainlp} is used to create a tokenizer with vocab size of 59,235 syllables.
    \item \textbf{SEFR tokenizer}; Stacked Ensemble Filter and Refine tokenizer (\textit{engine=best}) \cite{limkonchotiwat2020domain} based on probabilities from CNN-based \textit{deepcut} \cite{Kittinaradorn2019} with a vocab size of 92,177 words.
\end{itemize}

% trained on 15M sentences (Subject to be changed)%

\subsection{Train-Validation-Test Splits}

\paragraph{Assorted Thai Texts Dataset} After preprocessing and deduplication, we have a training set of 381,034,638 unique, mostly Thai sentences with sequence length of 5 to 300 words (78.5GB). The training set has a total of 16,957,775,412 words as tokenized by dictionary-based maximal matching \cite{pythainlp}, 8,680,485,067 subwords as tokenized by SentencePiece \cite{sentencepiece} tokenizer, and 53,035,823,287 characters.

\par We also randomly sampled 99,181 sentences (19.28MB) as validation set and 42,238,656 sentences (8GB) as test set. Both are preprocessed in the same manner as the training set.

\paragraph{Wikipedia-only Dataset} From \textit{Thai Wikipedia Dump}, we extract in a uniformly random manner 944,782 sentences for training set, 24,863 sentences for validation set and 24,862 sentences for test set.

% maybe a table for both datasets %

\subsection{Language Modeling}

\paragraph{Architecture}

We use the transformer \cite{vaswani2017attention} architecture of BERT (Base) (12 layers, 768 hidden dimensions, 12 attention heads) \cite{devlin2018bert}. Our setup is very similar to \cite{martin2019camembert} replacing BERT's WordPiece tokenizer with a SentencePiece tokenizer, with the exception of preprocessing rules applied before subword tokenization.

\paragraph{Pretraining Objective} We train the model with masked language modeling. To circumvent the word boundary issues in Thai, we opted to perform this at the subword level instead of whole-word level, even though the latter is reported to have better performance in English \cite{joshi2020spanbert}. In the same manner as BERT \cite{devlin2018bert} and RoBERTa \cite{liu1907roberta}, for each sequence, we sampled 15\% of the tokens and replace them with \foreignlanguage{thai}{<mask>} token. Out of the 15\%, 80\% is replaced with a \foreignlanguage{thai}{<mask>} token, 10\% is left unchanged and 10\% is replaced with a random token. The objective is to predict the tokens replaced with \foreignlanguage{thai}{<mask>} using cross entropy loss.

\paragraph{Pretraining} We pretrain $\text{RoBERTa}_{\textrm{BASE}}$ on both the \textit{Assorted Thai Texts dataset} and \textit{Wikipedia-only dataset}. The size of \textit{Wikipedia-only dataset} is about 0.57 GB which is comparatively low compared to the \textit{Assorted Thai Texts dataset}. Therefore, we manually tune the hyperparamters used for $\text{RoBERTa}_{\textrm{BASE}}$  pretraining for each training set in order to control the loss stability. The hyperparameters of the  $\text{RoBERTa}_{\textrm{BASE}}$  architecture and model pretraining are listed in Table \ref{tab:hyperpam_roberta-base_thwiki_and_78gb-text}.

%  Variation on type of tokens
\begin{table}[H]
    \centering
    \begin{tabular}{l c c}
         \hline
         \textbf{Hyperparameters}  & $\textbf{RoBERTa}_{\textrm{BASE}}$ (Wikipedia-only Dataset) & $\textbf{RoBERTa}_{\textrm{BASE}}$ (Assorted Thai Texts Dataset) \\ \hline
         
         Number of Layers & 12 & 12 \\ 
         Hidden size & 768 & 768 \\ 
         FFN hidden size & 3,072 & 3,072 \\ 
         Attention heads & 12 & 12 \\
         Dropout & 0.1 & 0.1 \\ 
         Attention dropout & 0.1 & 0.1 \\ 
         Max sequence length & 512 & 416 \\
         Effective batch size & 8,192 & 4,092 \\
         Warmup steps & 1,250 & 24,000 \\ 
         Peak learning rate & 7e-4 & 3e-4 \\
         Learning rate decay & Linear &  Linear \\
         Max steps &  31,250 &  500,000 \\ 
         Weight decay & 0.01 & 0.01 \\
         Adam $\epsilon$ & 1e-6 & 1e-6 \\
         Adam $\beta_1$ & 0.9 & 0.9 \\ 
         Adam $\beta_2$ & 0.98 & 0.999 \\
         FP16 training & True & True \\  \hline
         
    \end{tabular}
    \caption{Hyperparameters of  $\text{RoBERTa}_{\textrm{BASE}}$ used when pretrain on \textit{Assorted Thai Texts dataset} and \textit{Wikipedia-only dataset}.}
    \label{tab:hyperpam_roberta-base_thwiki_and_78gb-text}
\end{table}

\paragraph{WangchanBERTa} We name our pretrained language models according to their architectures, tokenizers and the datasets on which they are trained on. The models can be found on HuggingFace\footnote{https://huggingface.co/models}.

\begin{table}[hbt!]
\centering
% \arrayrulecolor[rgb]{1,1,1}
\begin{tabular}{llll} 
% \arrayrulecolor{black}
\hline
                          & Architecture & Dataset       & Tokenizer       \\ 
\hline
wangchanberta-base-wiki-spm   & RoBERTa-base & Wikipedia-only & SentencePiece   \\ 
% \arrayrulecolor[rgb]{1,1,1}
% \hline
wangchanberta-base-wiki-newmm & RoBERTa-base & Wikipedia-only & word (newmm)    \\ 
% \hline
wangchanberta-base-wiki-syllable   & RoBERTa-base & Wikipedia-only & syllable (newmm)  \\ 
% \hline
wangchanberta-base-wiki-sefr  & RoBERTa-base & Wikipedia-only & SEFR            \\ 
% \hline
wangchanberta-base-att-spm-uncased    & RoBERTa-base & Assorted Thai Texts  & SentencePiece   \\
% \arrayrulecolor{black}
\hline
\end{tabular}
\caption{WangchanBERTa model names}
\label{tab:model_names}
\end{table}

\section{Downstream Tasks}

We evaluate the downstream performance of our pretrained Thai $\text{RoBERTa}_{\textrm{BASE}}$ models on existing Thai sequence-classification and token-classification benchmark datasets. 

%table for datasets%

\subsection{Datasets}

We use train-valiation-test split as provided by each dataset as hosted on Huggingface Datasets.\footnote{https://huggingface.co/datasets} When not all splits are available, namely for \textit{Wongnai Reviews} and \textit{ThaiNER}, we sample respective splits in a uniformly random manner. The descriptive statistics of each datasets  are as follows:

\begin{table}[!hbt]
\centering
% \arrayrulecolor[rgb]{0,0,0,0}
\scalebox{0.765}{\begin{tabular}{llllcrrr} 
\toprule
Datasets & Label        & Style                           & Tasks                               & \multicolumn{1}{r}{Labels} & \multicolumn{1}{r}{Train}  & \multicolumn{1}{r}{Eval}  & \multicolumn{1}{r}{Test}  \\ 
% \arrayrulecolor{black}
\hline
wisesight\_sentiment     & category     & informal; social media posts    & multi-class sequence classification & 4                           & 21628                       & 2404                       & 2671                       \\ 
% \arrayrulecolor[rgb]{0,0,0,0}
\hline
wongnai\_reviews         & star\_rating & informal; restaurant reviews    & multi-class sequence classification & 5                           & \multicolumn{1}{r}{36000\textsuperscript{*}} & \multicolumn{1}{r}{4000\textsuperscript{*}} & 6203                       \\ 
\hline
generated\_reviews\_enth & review\_star & informal; product reviews       & multi-class sequence classification & 5                           & 141369                      & 15708                      & 17453                      \\ 
\hline
prachathai67k            & tags         & formal; news                    & multi-label sequence classification & 12                          & 54379                       & 6721                       & 6789                       \\ 
\hline
thainer                  & ner\_tags    & formal; news and other articles & token classification                & \multicolumn{1}{r}{13\textsuperscript{**}}   & \multicolumn{1}{r}{5079\textsuperscript{*}}  & \multicolumn{1}{r}{635\textsuperscript{*}}  & \multicolumn{1}{r}{621\textsuperscript{*,***}}  \\ 
\hline
lst20                    & pos\_tags    & formal; news and other articles & token classification                & 16                          & 67104                       & 6094                       & 5733                       \\ 
\hline
lst20                    & ner\_tags    & formal; news and other articles & token classification                & 10                          & 67104                       & 6094                       & 5733                       \\
\bottomrule
\end{tabular}}
% \arrayrulecolor{black}
\flushleft{\footnotesize{*Uniform randomly split with seed = 2020}}
\flushleft{\footnotesize{**We replace \textthai{B-ไม่ยืนยัน} and \textthai{I-ไม่ยืนยัน} which are extremely rare tags from ThaiNER with O}}
\flushleft{\footnotesize{***We removed examples on test set which did not fit in mBERT max token length to have a fair comparison among all models}}
\caption{Datasets for downstream tasks}
\label{tab:downstream_datasets}
\end{table}

\subsubsection{Sequence Classification}

\paragraph{Wisesight Sentiment} \cite{bact_2019_3457447} is a multi-class text classification dataset (sentiment analysis). The data are social media messages in Thailand collected from 2016 to early 2019. Each message is annotated as positive, neutral, negative, or question.

\paragraph{Wongnai Reviews} \cite{wongnai2018} is a multi-class text classification dataset (rating classification). The data are restaurant reviews and their respective ratings from 1 (worst) to 5 (best) stars.

\paragraph{Generated Reviews EN-TH} \cite{lowphansirikul2020scb} is a dataset that originally consists of product reviews generated by CTRL \cite{DBLP:journals/corr/abs-1909-05858} in English. It is translated to Thai as part of the \textit{scb-mt-en-th-2020} machine translation dataset. Translation is performed both by human annotators and models. We use only the translated Thai texts as a feature to predict review stars from 1 (worst) to 5 (best).

\paragraph{Prachathai-67k} is a multi-label text classification dataset (topic classification) based on news articles of Prachathai.com from August 24, 2004 to November 15, 2018 packaged by \cite{pythainlp}. We perform topic classification of the headline of each article, which can contain none to all of the following labels: politics, human rights, quality of life, international, social, environment, economics, culture, labor, national security, ict, and education.

\subsubsection{Token Classification}

\paragraph{ThaiNER v1.3} \cite{wannaphong2019} is a  6,456-sentence named entity recognition (NER) dataset created by expanding an unnamed,  2,258-sentence dataset by \cite{tirasaroj2012}. The NER tags are annotated by humans in IOB format.

\paragraph{LST20} \cite{boonkwan2020annotation} is a dataset with 5 layers of linguistic annotations: word boundaries, POS tagging, NER, clause boundaries, and sentence boundaries. NER tags are in IOBE format. We use the dataset for POS tagging and NER tasks.

\subsection{Benchmarking Models}

We provide benchmarks using traditional models (NBSVM for sequence classification and CRF for token classification), RNN-based models (ULMFit; only for sequence classification) and transformer-based models.% (as hosted on HuggingFace\footnote{https://huggingface.co/models}).

\paragraph{NBSVM} \cite{wang2012baselines} We adopt the NBSVM implementation by Jeremy Howard\footnote{https://www.kaggle.com/jhoward/nb-svm-strong-linear-baseline} as our strong baselines for sequence classification both multi-class and multi-label. The notable differences are substituting binarized ngram features with tf-idf features (uni- and bi-grams; minimum document frequency of 3, maximum document frequency of 90\%). We also apply the same cleaning rules as the language model, with the differences being adding repeated character tokens \textthai{<rep>} and repeated word tokens \textthai{<wrep>} instead of space tokens \textthai{<\_>}.

\par We perform hyperparameter tuning for penalty types (L1 and L2) and inverse of regularization strength (C=[1.0, 2.0, 3.0, 4.0]) and choose the models with the highest F1 scores (micro-averaged for multi-class and macro-averaged for multi-label classification). See Table \ref{tab:nbsvm_hyperparams}. For multi-label classifcation, we search for the best set of thresholds (ranging between 0.01 -- 0.99 with the step size of 0.01) that maximize macro-average F1-score on validation set.

\paragraph{ULMFit (thai2fit)} is an implementation of ULMFit language model finetuning for text classification \cite{howard2018universal}. \cite{charin_polpanumas_2021_4429691} pretrained a language model with vocab size of 60,005 words (tokenized by PyThaiNLP's \textit{newmm}) on \textit{Thai Wikipedia Dump}. We finetune the language model on the training set of each dataset for 5 epochs. Then that, we finetune for the sequence classification tasks using gradual unfreezing from the last one, two and three parameter groups with discriminative learning rates, for one epoch each. After that, we finetune all the weights of the model for 5 epochs. The checkpoints with the highest accuracy scores (validation losses for multi-label classification) are chosen to perform on the test sets. See Table \ref{tab:thai2fit_hyperparams}. Lastly, we search for the best set of thresholds (ranging between 0.01 -- 0.99 with the step size of 0.01) that maximize macro-average F1-score on validation set.

\paragraph{Conditional Random Fields (CRF)} \cite{lafferty2001conditional} We use the CRFSuite implementation \cite{okazaki2007crfsuite} of conditional random fields as a strong baseline for POS and NER tagging tasks. We generate the features by extracting unigrams, bigrams and trigrams features within a sliding window of three timesteps, before and after the current token (beginning and ending of sentences are padded with \textit{xxpad} tokens). We finetune L1 and L2 penalty combinations using 10,000 randomly sampled sentences for \textit{LST20} and the entire training set for \textit{ThaiNER}. With hyperparameters with the best F1 score (micro-averaged) on the validation set, we train on the entire training sets and report performances on the test sets. See Table \ref{tab:crf_hyperparams}. We run each CRF model for 500 iterations.

\paragraph{Transformer-based models} We use the same finetuning scheme for all transformer-based models, namely XLM-RoBERTa-base \cite{conneau2019unsupervised}, BERT-base-multilingual-cased \cite{DBLP:journals/corr/abs-1810-04805}, wangchanberta-base-wiki-tokenizer (\textit{spm}, \textit{newmm}, \textit{syllable}, \textit{sefr}), and wangchanberta-base-att-spm-uncased.  For the sequence classification task,
we preprocess each dataset with the rules described in \ref{subsection:text_preprocessing}. We then finetune each pretrained language model on downstream tasks for 3 epochs. The criteria to select the best epoch is the validation micro-average F1-score for multi-class classification and macro-average F1-score for multi-label classification.
The batch size is set to 16. The  The learning rate is warmed up over the first 10\% of  steps to the value of 3e-5 and linearly decayed to zero. We finetune models with FP16 mixed precision training. All models are optimized with Adam \cite{kingma2014adam} ($\beta_1 =0.9$, $\beta_2 = 0.999$, $\epsilon = $1e-8, $L_2$ weight decay $=$ 0.01) with corrected bias. For multi-label classification head, we search for the best set of thresholds (ranging between 0.01 -- 0.99 with the step size of 0.01) that maximize macro-average F1-score on validation set.
%table for finetuning parameters%

% Hyperparam for Token CLS
\par For the token classification tasks, we finetune each pretrained language models for 6 epochs. The criteria to select the best epoch is the validation loss. The batch size is set to 32. The learning rate is warmed up
over the first 10\% of  steps to the value of 3e-5 and linearly decayed to zero. We finetune models with FP16 mixed precision training. All models are optimized with Adam with the parameters as same as the sequence classification task. 

%Each pretrained model is attached with a classification head comprised of 1 linear layer (768 hidden units), a dropout layer with the rate of 0.1, and output layer.
\section{Results}
\subsection{Language Modeling} 
\par The following table shows the performance   $\text{RoBERTa}_{\textrm{BASE}}$ trained on \textit{Wikipedia-only dataset}. There are four variations of tokenization including subword-level with SentencePiece \cite{sentencepiece}, word-level and syllable-level with PyThaiNLP \cite{pythainlp} tokenizer (denoted as \textit{newmm} and \textit{syllable} respectively), and  stacked-ensemble, word-level tokenizer \textit{sefr} \cite{limkonchotiwat2020domain}.

\begin{table}[H]
    % \centering
    \scalebox{0.9}{\begin{tabular}{ l c c c c }
         \hline
         
         \multirow{2}{*}{Model Name} & \multirow{2}{*}{Vocab Size} & \multirow{2}{*}{Number of Training Examples} & \multicolumn{2}{c}{Best Checkpoint} \\ 
         \cmidrule(lr){4-5} & & &  Validation loss & Steps \\
         \hline
         \multicolumn{5}{l}{\textit{Pretraining on Wikipedia-only dataset}:} \\
         \addlinespace[0.075cm]
         \hspace{0.25cm}wangchanberta-base-wiki-spm & 24,000 & 116,715 & 1.5127 & 7,000 \\ 
         \hspace{0.25cm}wangchanberta-base-wiki-newmm & 97,982  & 119,074 & 1.4990 &  5,000 \\ 
         \hspace{0.25cm}wangchanberta-base-wiki-syllable & 59,235  & 167,279 & 0.8068 & 8,000 \\ 
         \hspace{0.25cm}wangchanberta-base-wiki-sefr & 92,177 & 125,177 & 1.2995 & 4,500 \\
         \addlinespace[0.15cm]
         \multicolumn{5}{l}{\textit{Pretraining on Assorted Thai Texts dataset} (Currently, the model has not reached the max steps):} \\
        
         \multirow{2}{*}{\hspace{0.25cm}wangchanberta-base-att-spm-uncased} & 
         \multirow{2}{*}{25,000} & 
         \multirow{2}{*}{382 M} & 
         \multirow{2}{*}{2.551} &  360,000 \\
         & & & & \footnotesize{(latest checkpoint)} \\ 
         \hline
    \end{tabular}}
    \caption{The vocab size, number of training examples, and best checkpoint of the $\text{RoBERTa}_{\textrm{BASE}}$ models trained on Thai Wikipedia corpus for each type of input tokens and Assorted Thai Texts dataset.}
    \label{tab:thwiki_roberta_based}
\end{table}

For the $\text{RoBERTa}_{\textrm{BASE}}$  trained on \textit{Assorted Thai Texts dataset}, we only trained with subword token built with SentencePiece \cite{sentencepiece} due to the limited computational resources. 

\subsection{Downstream Tasks}

We choose models to perform on the test set based on their performance on the validation sets. For multi-class sequence classification and token classification, we optimize our models for the highest micro-averaged F1 score. For multi-label sequence classification, we optimize for the highest macro-averaged F1 score, as it is less affected by class imbalance. Moreover, for multi-label sequence classification, we also find the best probability threshold for each label based on the validation set. We report the performance of these optimized models on the test sets.

For sequence classification tasks, our model trained on the \textit{Assorted Thai Texts dataset} outperfroms both strong baselines and other transformer-based architecture on all downstream tasks except Generated Reviews (EN-TH). This may be attributed to the fact that the dataset is translated from generated texts in English, thus multi-lingual pretraining of XLMR gives it the advantage. \ref{tab:downstream_task_results-sequence_classification}.

For token classification tasks, our model trained on the \textit{Assorted Thai Texts dataset} achieves the highest micro-averaged F1 score in all tasks except POS tagging in \textit{ThaiNER} dataset. This could be attributed to the fact that the POS tags in \textit{ThaiNER} are machine-generated and thus more suited for the baseline model CRF. See Table \ref{tab:downstream_task_results-token_classification}.
% downstream result table %

\begin{table}[H]
\renewcommand*{\arraystretch}{1.0}
\begin{tabular}{ l c c c  c }
\hline
\addlinespace[0.1cm]
\multirow{2}*{Model} & 
\multirow{2}*{\text{Wisesight}} & 
\multirow{2}*{\text{Wongnai}}  & \text{Generated Reviews \scriptsize{(EN--TH)}}  & \multirow{2}*{\text{Prachathai}} \\
& & & (Review rating) & \\
\addlinespace[0.1cm]
\hline 
\addlinespace[0.08cm]
\multicolumn{5}{l}{\textit{Existing multilingual models}:}\\
\addlinespace[0.075cm]
\hspace{0.25cm} XLMR \cite{conneau2019unsupervised}  & 73.57 / 62.21 &  62.57 / 52.75  & \textbf{64.91} / \textbf{60.29}       & 68.18 / 63.14 \\
\hspace{0.25cm} mBERT \cite{devlin2018bert}          & 70.05 / 57.81 &  47.99 / 12.97  & 62.14 / 57.20        & 66.47 / 60.11 \\
\addlinespace[0.125cm]
\multicolumn{5}{l}{\textit{Our baseline models}:} \\ 
\addlinespace[0.075cm]
\hspace{0.25cm} Naive Bayes SVM                      & 72.03 / 54.67   & 58.38 / 39.75   & 59.68 / 52.17        & 66.77 / 60.73 \\ 
\hspace{0.25cm} ULMFit (thai2fit)                    & 70.95 / 60.62   & 61.79 / 48.04   & 64.33 / 59.33        & 66.21 / 60.21 \\  
\addlinespace[0.125cm]\multicolumn{5}{l}{\textit{Our pretrained} $\text{RoBERTa}_{\textrm{BASE}}$ \textit{models}:} \\
\addlinespace[0.075cm]
\hspace{0.25cm} wangchanberta-base-wiki-spm         & 73.94 / 60.13    &  60.60  / 48.17  & 63.43 / 58.43       & 68.85 / 63.46 \\
\hspace{0.25cm} wangchanberta-base-wiki-newmm       & 72.74 / 55.87    &  59.81  / 45.75  & 63.70 / 58.41       & 68.78 / 63.50 \\
\hspace{0.25cm} wangchanberta-base-wiki-syllable         & 73.42 / 59.12    &  60.36 / 46.68   & 63.53 / 58.73       & 68.90 / 63.59 \\
\hspace{0.25cm} wangchanberta-base-wiki-sefr        & 70.80 / 59.51    &  59.83 / 48.21   & 63.31 / 58.85  & 67.45 / 61.14 \\
\hspace{0.25cm} wangchanberta-base-att-spm-uncased  & \textbf{76.19} / \textbf{67.05}    &  \textbf{63.05} / \textbf{52.19}   & 64.66 / 59.54       & \textbf{69.78} / \textbf{64.90} \\ 
\addlinespace[0.06cm]
\hline
\end{tabular}
\caption{Test set results for RoBERTa BASE models we pretrain and existing multilingual language models inclduing XLM $\text{RoBERTa}_{\textrm{BASE}}$ (XLMR) and Multilingual $\text{BERT}_{\textrm{BASE}}$ (mBERT). The metrics we report are micro-average and macro-average F1 score. }
\label{tab:downstream_task_results-sequence_classification} 
\end{table}

\begin{table}[H]
    % \centering
    \renewcommand*{\arraystretch}{1.0}
    \begin{tabular}{ l  c c c  }
         \hline
         \addlinespace[0.1cm]
         \multirow{2}*{Model} & ThaiNER & \multicolumn{2}{c}{LST20}
         \\ \cmidrule(lr){2-2} \cmidrule(lr){3-4}
          & NER &  POS & NER \\
         \addlinespace[0.1cm]
         \hline 
         \addlinespace[0.08cm]
         \multicolumn{4}{l}{\textit{Existing multilingual models}:} \\ 
         \addlinespace[0.075cm]
         \hspace{0.25cm} XLMR \cite{conneau2019unsupervised} & 83.25 / 67.23                & 96.57 / 85.00              & 73.61 / 68.67 \\
         \hspace{0.25cm} mBERT \cite{devlin2018bert}         & 81.48 /  73.97               & 96.44 / \textbf{85.86}        & 75.05  / 68.25 \\
         \addlinespace[0.125cm]
         \multicolumn{4}{l}{\textit{Our baseline models}:} \\ 
         \addlinespace[0.075cm]
         \hspace{0.25cm} Conditional Random Fields (CRF)     & 78.98 /  \textbf{81.85}      & 96.28 / 81.28      & 75.94 / 72.13  \\
         \addlinespace[0.125cm]
         \multicolumn{4}{l}{\textit{Our pretrained} $\text{RoBERTa}_{\textrm{BASE}}$ \textit{models}:} \\
         \addlinespace[0.075cm] 
         \hspace{0.25cm} wangchanberta-base-wiki-spm         & 56.64 / 55.34     & 96.18  / 83.99        & 77.12 / 71.32    \\
         \hspace{0.25cm} wangchanberta-base-wiki-newmm       & 58.54 / 47.71     & 96.14  / 83.11        & 76.59 / 70.57 \\
         \hspace{0.25cm} wangchanberta-base-wiki-syllable         & 83.23 / 76.64     & 96.06  / 83.98        & 76.45 / 70.37\\
         \hspace{0.25cm} wangchanberta-base-wiki-sefr        & 85.04 / 77.73     & 96.36 / 85.24         & 76.25 / 69.34\\
         \hspace{0.25cm} wangchanberta-base-att-spm-uncased  & \textbf{86.49} / 79.29  & \textbf{96.62} / 85.44             & \textbf{78.01} / \textbf{72.25}   \\ 
         \addlinespace[0.06cm]
         \hline
        
    \end{tabular}
    \caption{Test set results for RoBERTa BASE models we pretrain and existing multilingual language models inclduing XLM $\text{RoBERTa}_{\textrm{BASE}}$ (XLMR) and Multilingual $\text{BERT}_{\textrm{BASE}}$ (mBERT). The metrics we report are micro-average and macro-average F1 score. }
    \label{tab:downstream_task_results-token_classification} 
\end{table}
\section{Discussions and Future Works}

\par Consistent with previous works on language modeling, we found that training on large datasets such as our \textit{Assorted Thai Texts dataset} yield better downstream performance. The only case when a multi-lingual model (XLMR) outperforms our largest mono-lingual model is when the training data include multi-lingual elements namely the English-to-Thai translated texts of \textit{Generated Reviews EN-TH}. From our experiments on the \textit{Wikipedia-only dataset}, we did not find any notable diferrence in downstream performance for sequence classification or token classification tasks.

\par Another area we will explore in the future is the inherent biases on our relatively large language models. Previous works including \cite{sheng2019woman} \cite{nadeem2020stereoset} \cite{nangia2020crowspairs} have detected social biases within large language models trained in English. Our next step in this direction is to create similar bias-measuring datasets in Thai contexts to detect the biases in our language models.

\par We pretrain our language models on publicly available datasets. Two main concerns that have been raised about similar models are copyrights and privacy. All datasets used to train our models are based on publicly available data. Publicly available social media data are packaged and provided to use by Wisesight\footnote{https://wisesight.com} (\textit{wisesight-large}) and Chaos Theory\footnote{https://www.facebook.com/ChaosTheoryCompany/} (\textit{pantip-large}). Unless specified otherwise in the distribution of datasets, all rights belong to the content creators. We provide the weights of our pretrained language models under CC-BY-SA 4.0. Our models are trained as feature extractors for downstream tasks, and not generative tasks. Reproduction of training data can happen \cite{carlini2020extracting} albeit at much lower chance than language models trained specifically for generative tasks.
\section{Acknowledgements}

We thank Wisesight\textsuperscript{16}, Chaos Theory\textsuperscript{17} and Pantip.com for providing what has become, to the best of our knowledge, the largest and most diverse high-quality training data in Thai for language modeling.
% \selectlanguage{english}

\bibliographystyle{apalike}  
\bibliography{references}

\newpage
\section{Appendix}

\begin{table}[hbt!]
\begin{tabular}{llrr}
% \arrayrulecolor[rgb]{0,0,0}
\toprule
datasets{[}features:labels{]}                                                   & penalty & \multicolumn{1}{l}{C} & \multicolumn{1}{l}{f1} \\
\hline
\multirow{8}{*}{wisesight\_sentiment{[}texts:category{]}}                       & l2      & 3                     & 0.720466               \\
 & l2 & 2 & 0.718386 \\
 & l2 & 4 & 0.715474 \\
 & l1 & 2 & 0.710067 \\
 & l1 & 3 & 0.707571 \\
 & l2 & 1 & 0.707571 \\
 & l1 & 1 & 0.706323 \\
 & l1 & 4 & 0.705075 \\
 \hline
\multirow{8}{*}{wongnai\_reviews{[}review\_body:star\_rating{]}}                & l1      & 2                     & 0.57125                \\
 & l2 & 4 & 0.5705   \\
 & l2 & 3 & 0.56675  \\
 & l1 & 3 & 0.5635   \\
 & l2 & 2 & 0.5605   \\
 & l1 & 4 & 0.55425  \\
 & l1 & 1 & 0.55325  \\
 & l2 & 1 & 0.54     \\
 \hline
\multirow{8}{*}{generated\_reviews\_enth{[}translation{[}th{]}:review\_star{]}} & l2      & 2                     & 0.593265               \\
 & l2 & 3 & 0.591609 \\
 & l2 & 1 & 0.590718 \\
 & l2 & 4 & 0.5904   \\
 & l1 & 2 & 0.590209 \\
 & l1 & 1 & 0.590018 \\
 & l1 & 3 & 0.58467  \\
 & l1 & 4 & 0.577222 \\
 \hline
\multirow{8}{*}{prachathai67k{[}title:multilabel{]}}                            & l2      & 1                     & 0.61105                \\
 & l2 & 2 & 0.607425 \\
 & l2 & 3 & 0.60561  \\
 & l2 & 4 & 0.601663 \\
 & l1 & 1 & 0.59017  \\
 & l1 & 2 & 0.585137 \\
 & l1 & 3 & 0.578731 \\
 & l1 & 4 & 0.574738 \\
 \bottomrule
\end{tabular}
\caption{NBSVM Hyperparameter Tuning Results}
\label{tab:nbsvm_hyperparams}
\end{table}

\begin{table}[hbt!]
\centering
% \arrayrulecolor[rgb]{0,0,0}
\scalebox{0.66}{\begin{tabular}{lllrrrr} 
\toprule
datasets[features:labels]                                                & finetuning & unfreezing & \multicolumn{1}{l}{epoch} & \multicolumn{1}{l}{train\_loss} & \multicolumn{1}{l}{valid\_loss} & \multicolumn{1}{l}{accuracy}  \\ 
\hline
\multirow{13}{*}{wisesight\_sentiment[texts:category]}                   & LM         & all        & 0                         & 4.459779                        & 4.256248                        & 0.330593                      \\
                                                                         & LM         & all        & 1                         & 4.261896                        & 4.081441                        & 0.348543                      \\
                                                                         & LM         & all        & 2                         & 4.042319                        & 3.979969                        & 0.358797                      \\
                                                                         & LM         & all        & 3                         & 3.854878                        & 3.939566                        & 0.364824                      \\
                                                                         & LM         & all        & 4                         & 3.754257                        & 3.932823                        & 0.365743                      \\
                                                                         & CLS        & last 1     & 0                         & 0.90485                         & 0.800313                        & 0.653078                      \\
                                                                         & CLS        & last 2     & 0                         & 0.834293                        & 0.762427                        & 0.675957                      \\
\bottomrule
\end{tabular}}

\flushleft{finetuning LM: languagel model, CLS: classification; unfreezing all: all layers, last X: last X layer groups}
\caption{ULMFit (thai2fit) Hyperparameter Tuning Results}
\label{tab:thai2fit_hyperparams}
\vspace{-4.5pt}
\end{table}
\begin{table}[hbt!]
% \arrayrulecolor[rgb]{0,0,0}
\begin{tabular}{lrrrr}
\toprule
datasets{[}features:labels{]} & \multicolumn{1}{l}{l1 penalty} & \multicolumn{1}{l}{l2 penalty} & \multicolumn{1}{l}{f1\_micro} & \multicolumn{1}{l}{f1\_macro} \\
\hline
\multirow{9}{*}{lst20{[}tokens:ner\_tags{]}}   & 0.5 & 0   & 0.717296 & 0.627721 \\
                                               & 1   & 0   & 0.716445 & 0.622451 \\
                                               & 0   & 0   & 0.688666 & 0.615289 \\
                                               & 0   & 0.5 & 0.703625 & 0.602803 \\
                                               & 0.5 & 0.5 & 0.699979 & 0.590872 \\
                                               & 0   & 1   & 0.694041 & 0.586303 \\
                                               & 1   & 0.5 & 0.692479 & 0.585022 \\
\bottomrule
\end{tabular}
\caption{CRF Hyperparameter Tuning Results}
\label{tab:crf_hyperparams}
\end{table}

%% TODO CLS REport for NER POS (POS left, NER right )`

%% 1, CRF POS, NER 
\begin{table}[ht]

%
 
\caption{ CRF -- per-class precision, recall and F1-score on test set of ThaiNER }
\label{tab:classification_report-thainer-crf_best}
\end{table}

\begin{table}[ht]
\renewcommand*{\arraystretch}{1.0}
%
 
\caption{ CRF -- per-class precision, recall and F1-score on test set of LST20 }
\label{tab:classification_report-lst20-crf_best}
\end{table}

%% 2. roberta-thwiki-spm

\begin{table}[ht]
\renewcommand*{\arraystretch}{1.0}
%
 
\caption{ wangchanberta-thwiki-spm -- per-class precision, recall and F1-score on test set of ThaiNER }
\label{tab:classification_report-thainer-wangchanberta-thwiki-spm}
\end{table}

\begin{table}[ht]
\renewcommand*{\arraystretch}{1.0}
%
 
\caption{ wangchanberta-thwiki-spm -- per-class precision, recall and F1-score on test set of LST20 }
\label{tab:classification_report-lst20-wangchanberta-thwiki-spm}
\end{table}

%% 3. roberta-thwiki-newmm

\begin{table}[ht]
\renewcommand*{\arraystretch}{1.0}
%
 
\caption{ wangchanberta-thwiki-newmm  -- per-class precision, recall and F1-score on test set of ThaiNER }
\label{tab:classification_report-thainer-wangchanberta-thwiki-newmm}
\end{table}

\begin{table}[ht]
\renewcommand*{\arraystretch}{1.0}
%
 
\caption{ wangchanberta-thwiki-newmm -- per-class precision, recall and F1-score on test set of LST20 }
\label{tab:classification_report-lst20-wangchanberta-thwiki-newmm}
\end{table}

%%%%%%%%%%%%%%%%%%%%%%%%%%%%%%%%%%%%%%%%%%%%%%%%%%%%%%%%%%%%%%%%%%%%%%%%%%%%%%%%%%%%%%%%%%%%%%%%%%%%%%%%%%%%%%%%%%
%%%%%%%%%%%%%%%%%%%%%%%%%%%%%%%%%%%%%%%%%%%%%%%%%%%%%%%%%%%%%%%%%%%%%%%%%%%%%%%%%%%%%%%%%%%%%%%%%%%%%%%%%%%%%%%%%%

%% 4. roberta-thwiki-syllable

\begin{table}[ht]
\renewcommand*{\arraystretch}{1.0}
%
 
\caption{ wangchanberta-thwiki-syllable -- per-class precision, recall and F1-score on test set of ThaiNER }
\label{tab:classification_report-thainer-wangchanberta-thwiki-syllable }
\end{table}

\begin{table}[ht]
\renewcommand*{\arraystretch}{1.0}
%
 
\caption{ wangchanberta-thwiki-syllable -- per-class precision, recall and F1-score on test set of LST20 }
\label{tab:classification_report-lst20-wangchanberta-thwiki-syllable }
\end{table}

%% 5. roberta-thwiki-sefr

\begin{table}[ht]
\renewcommand*{\arraystretch}{1.0}

%
 
\caption{ wangchanberta-thwiki-sefr -- per-class precision, recall and F1-score on test set of ThaiNER }
\label{tab:classification_report-thainer-wangchanberta-thwiki-sefr}
\end{table}

\begin{table}[ht]
\renewcommand*{\arraystretch}{1.0}

%
 
\caption{ wangchanberta-thwiki-sefr -- per-class precision, recall and F1-score on test set of LST20 }
\label{tab:classification_report-lst20-wangchanberta-thwiki-sefr}
\end{table}

%% 6. roberta-78g-spm-uncased

\begin{table}[ht]
\renewcommand*{\arraystretch}{1.0}

%
 
\caption{ wanchanberta-base-att-spm-uncased -- per-class precision, recall and F1-score on test set of ThaiNER }
\label{tab:classification_report-thainer-wanchanberta-base-att-spm-uncased }
\end{table}

\begin{table}[ht]
\renewcommand*{\arraystretch}{1.0}

%
 
\caption{ wanchanberta-base-att-spm-uncased -- per-class precision, recall and F1-score on test set of LST20 }
\label{tab:classification_report-lst20-wanchanberta-base-att-spm-uncased }
\end{table}

%% 7. mbert

\begin{table}[ht]
\renewcommand*{\arraystretch}{1.0}

%
 
\caption{ mBERT -- per-class precision, recall and F1-score on test set of ThaiNER }
\label{tab:classification_report-thainer-mbert}
\end{table}

\begin{table}[ht]
\renewcommand*{\arraystretch}{1.0}

%
 
\caption{ mBERT -- per-class precision, recall and F1-score on test set of LST20 }
\label{tab:classification_report-lst20-mbert}
\end{table}

%% 8. xmlr

\begin{table}[ht]
\renewcommand*{\arraystretch}{1.0}

%
 
\caption{ XLMR -- per-class precision, recall and F1-score on test set of ThaiNER }
\label{tab:classification_report-thainer-xlmr}
\end{table}

\begin{table}[ht]
\renewcommand*{\arraystretch}{1.0}

%
 \\
\end{tabular} 
\caption{ XLMR -- per-class precision, recall and F1-score on test set of LST20 }
\label{tab:classification_report-lst20-xlmr}
\end{table}

\end{document}